\pdfoutput=1

\documentclass[11pt]{article}
\usepackage[preprint]{acl}
\usepackage{graphicx}
\usepackage{multirow}
\usepackage{pifont}
\usepackage{times}
\usepackage{caption}
\usepackage{subcaption}
\usepackage{pgfplots}
\pgfplotsset{compat=1.18}
\usepackage{tikz}
\usetikzlibrary{patterns}
\usepackage[T1]{fontenc}
\usepackage{inconsolata}
\usepackage{tabularx}
\usepackage{array}
\usepgfplotslibrary{groupplots}
\usepackage{colortbl}
\usepackage{tcolorbox}
\tcbuselibrary{theorems}
\usepackage{cleveref}
\newtcbtheorem[]{mytheo}{Finding}%
{colback=green!5,colframe=green!35!black,fonttitle=\bfseries}{th}
\crefformat{section}{\S#2#1#3}

\title{Exploring Robustness of Multilingual LLMs on Real-World Noisy Data}

\author{Amirhossein Aliakbarzadeh\textsuperscript{$\clubsuit$} \and Lucie Flek\textsuperscript{$\clubsuit\spadesuit$} \and Akbar Karimi\textsuperscript{$\clubsuit\spadesuit$}\\
\vspace{-1mm}\\
\textsuperscript{$\clubsuit$}Conversational AI and Social Analytics (CAISA) Lab, University of Bonn, Germany\\
\textsuperscript{$\spadesuit$}Lamarr Institute for Machine Learning and Artificial Intelligence, Germany\\
\texttt{amir.akbarzadeh95@gmail.com}\\
\texttt{\{flek, ak\}@bit.uni-bonn.de}}

\newcolumntype{Y}{>{\centering\arraybackslash}X}
\newcolumntype{m}{>{\hsize=.05\hsize}X}

\begin{document}
\maketitle
\begin{abstract}
Large Language Models (LLMs) are trained on Web data that might contain spelling errors made by humans. But do they become robust to similar real-world noise? In this paper, we investigate the effect of real-world spelling mistakes on the performance of 9 language models, with parameters ranging from 0.2B to 13B, in 3 different NLP tasks, namely Natural Language Inference (NLI), Name Entity Recognition (NER), and Intent Classification (IC). We perform our experiments on 6 different languages and build a dictionary of real-world noise for them using the Wikipedia edit history. We show that the performance gap of the studied models on the clean and noisy test data averaged across all the datasets and languages ranges from \boldmath{$2.3$} to \boldmath{$4.3$} absolute percentage points. In addition, mT5 models, in general, show more robustness compared to BLOOM, Falcon, and BERT-like models. In particular, mT5 (13B), was the most robust on average overall, across the 3 tasks, and in 4 of the 6 languages\footnote{\url{https://github.com/caisa-lab/LLMs-Real-World-Noise-Robustness}}. 
\end{abstract}

\section{Introduction}

\begin{figure}
    \centering
    \includegraphics[width=0.95\linewidth]{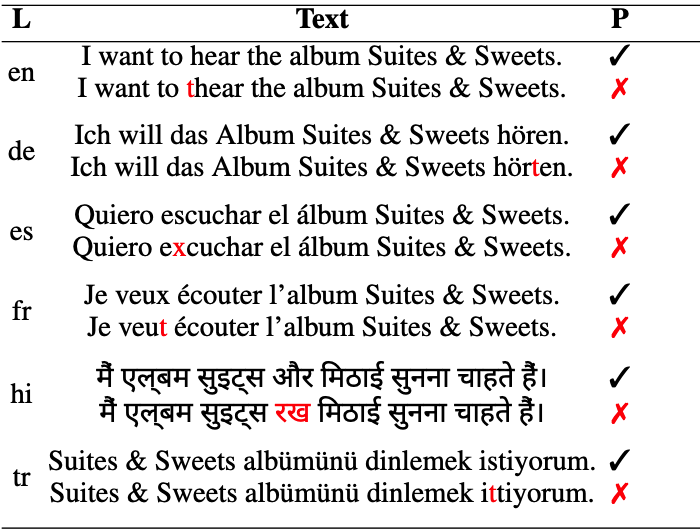}
    \caption{Typos that users make may lead LLMs to misclassify the input sentences. The sentences in the table are clean and noisy test samples of the intent classification data (SNIPS) that were misclassified by the studied models after typical typographical errors from our WikiTypo corpus were inserted.}
    \label{fig:example}
\end{figure}

Multilingual Large Language Models (LLMs) such as mT5 \cite{xue2020mt5} and BLOOM \cite{le2022bloom} have shown remarkable performance across a variety of tasks and languages. Given that they are usually trained on Web data, they might have already seen noisy data such as words with spelling errors during their pre-training. Figure \ref{fig:example} shows examples of such errors in different languages. It also shows that these errors can negatively influence the model prediction. This raises the critical question of how well these models perform in the presence of real-world noise. Additionally, given the multilingual nature of LLM interactions, understanding how performance varies across different languages is essential. Finally, the wide range of LLM sizes, from millions to billions of parameters, prompts the question of whether larger models are inherently more robust to real-world noise than smaller ones.

Previous research has only partially addressed these questions, often focusing on monolingual datasets or using simulated noise that may not accurately reflect real-world errors \cite{robustness_to_input_perturbations, wang2023large}.

A closely related study by \citet{stickland2023robustification} investigated real-world noise in a multilingual setting using mBERT \cite{devlin2018bert} and XLM-R \cite{conneau2019unsupervised}. However, given the varying behaviors of language models at different scales \cite{du2023shortcut}, it is crucial to examine the robustness of larger models to real-world noise. Furthermore, limitations in data accessibility due to corporate policies led us to create our own noisy test sets, which will be publicly available upon publication.


Our research aims to address these gaps by constructing a collection of real-world noisy data, called WikiTypo, using Wikipedia edit history. Specifically, we ask three research questions (RQ): 
\begin{itemize}
    \item \textbf{RQ1: } Are larger models more robust to real-world noisy data than smaller models?
    \item \textbf{RQ2: } Are different tasks equally sensitive to real-world noise?
    \item \textbf{RQ3: } How does the model performance differ from English to other languages under the noise?
\end{itemize}

We evaluate the robustness of multilingual LLMs against real-world noisy inputs by fine-tuning them on multilingual datasets and assessing their performance on clean and noisy test sets across three NLP tasks: Natural Language Inference (NLI), Name Entity Recognition (NER), and Intent Classification (IC). To create a noisy NER test set, we employ a word augmenter from the NLPAug library \cite{ma2019nlpaug}. 

Our findings reveal that all studied models exhibit vulnerability to real-world noise, but the performance gaps vary across tasks (\cref{sec:results}). NLI tasks generally show the largest gaps, while intent classification demonstrates the smallest (\cref{sub:tasks}). Moreover, decoder-only models like BLOOM and Falcon tend to exhibit larger gaps in the NER task (\cref{sub:models}).

\noindent
Our contributions are twofold: 1) We build a multilingual dictionary of real-world typos (called WikiTypo) for five languages from Wikipedia and create noisy data for three different NLP tasks, namely NLI, NER, and Intent Classification (IC); 2) We evaluate the robustness of nine multilingual language models varied in size from a fraction of a billion parameters to 13 billion parameters on WikiTypo. 

\section{Related Work}
Given the widely used devices that allow user interactions in the form of written textual communication with language models, several studies have investigated the effects of user errors on these models. Some prior works have studied language models in the context of noisy test data that were created artificially using simulation techniques \cite{srivastava2020noisy, cai2022context, almagro2023lea}. 
To simulate the effects of typos and misspelling errors, the authors inserted noise by randomly selecting a percentage of characters in close proximity on the QWERTY keyboard layout to mimic errors in data.
However, real-world applications will frequently face authentic noise, including linguistic variations which are absent in many benchmarks \cite{stickland2023robustification}.

Moreover, some works have focused on monolingual datasets (mainly English) and failed to explore the capacity of multilingual LLMs using multilingual test sets  \cite{srivastava2020noisy, robustness_to_input_perturbations, wang2023large}. These studies on textual noise have been limited to the exploration of noise within English language datasets. Although large multilingual models perform impressively on various tasks and languages, their performance is usually degraded in non-English languages, especially low-resource ones \cite{etxaniz2023multilingual}. 



Finally, various studies have mostly focused on BERT-like models thus far, indicating the need for studying larger language models that have gained popularity and widespread usage recently \cite{naplava2021understanding,xu2021robust, stickland2023robustification}. Prior studies have demonstrated the robustness of models such as BERT, XLM-Roberta, and XLNet to textual noise. Notably, these models achieve impressive performance despite having parameter sizes below 0.3 billion. This observation highlights a significant disparity between the models evaluated in prior research and the current LLMs. The parameter size of contemporary LLMs often has billions of parameters, suggesting a potential for further analysis of their capabilities against noise.


Filling the gap between multilingual models, model size, and real-world noise, we evaluate the robustness of 6 language models (with 0.2B to 13B parameters) to real-world multilingual noisy data in 6 languages for 3 NLP tasks.

\section{WikiTypo: A Collection of Real-World Typos from Wikipedia}

Wikipedia is an online encyclopedia that anyone can have access to edit the articles. A massive group of volunteers, called Wikipedians, keep it running by adding and updating information. 
Anyone can access the content of Wikipedia articles and the history of each page. This history shows the order in which the edits were made, and how the content changed between versions. This timeline, therefore, is called revision or edit history.

While some projects have focused on finding typos in Wikipedia edit history, there is a lack of publicly available code to do this for various languages. Similar to \citet{tanaka2020building} who work on mining typos from Wikipedia for Japanese articles, and our baseline \cite{stickland2023robustification}, we have leveraged Wikipedia revisions on random pages to find typos for specific languages. 

To identify the typos and their corrections, first, we consider each article and its revisions and parse the main pages using the BeatutifulSoap library\footnote{\url{https://pypi.org/project/beautifulsoup4}}. Then, we filter out the added and removed words on each page and extract a pair of words that 1) have one character-level Levenshtein edit distance from each other; 2) do not contain any number of special characters; and 3) have at least two characters each. The result is a dictionary of misspelled words with their corresponding correct spelling. Table \ref{tab:typos_number} shows the number of entries in our dictionary for each language.

\begin{table}[t]
    \centering
    \small
    \setlength{\tabcolsep}{7pt}
    \begin{tabular}{l r}
        \hline
        \textbf{Languages} & \textbf{Number of typos}\\
        \hline
        English (en) & 9370\\
        German (de) & 15000\\
        Spanish (es) & 8200\\
        French (fr) & 14900\\
        Hindi (hi) & 3300\\
        Turkish (tr) & 4060\\
        \hline
    \end{tabular}
    
    \caption{Number of typos for each language collected after inspecting 320000 pages.}
    \label{tab:typos_number}
    
\end{table}

\noindent
\textbf{NLPAug augmenter} \cite{ma2019nlpaug}.
NLPAug is a text data augmentation tool that handles various languages. In particular, its keyboard augmenter module can be used to inject typos into text data, mimicking the types of errors people commonly make when typing. For instance, it might replace an \texttt{\Large i} with an \texttt{\Large o} or vice versa, considering the proximity of these keys on a keyboard (based on Levenshtein distance).

\noindent
\textbf{Noisy test data creation.}
The WikiTypo noise dictionary is used to construct the noisy test sets for XNLI and SNIPS datasets. We use NLPAug augmenter to create noisy data for the WikiANN dataset since some sentences, in this dataset, contain only 2 or 3 words, and most of the words are proper nouns such as person, location, or organization names. In this case, the use of WikeTypo is limited since in Wikipedia edit history, misspelled words are rarely found.

To create a noisy version of the original test set (for all datasets), we randomly replace words in a sentence with their incorrect version from the noise dictionary. The maximum number of augmented words ($m$) and the ratio of the sentence ($r$) to be changed can be set beforehand as the hyperparameters of the noise insertion procedure. We choose $r=0.2$ and $m=4$ as the default values resulting in the noisy test sets with details in Table \ref{tab:noise_stat}.

\begin{table}[t]
    \centering
    \small
    \setlength{\tabcolsep}{3pt}
    
    \begin{tabular}{l | c| c| c| c| c}
    \hline
    & &  \textbf{\#Tokens} & \textbf{Avg \#Tokens} & \textbf{\#Noise} & \textbf{Noise Ratio} \\
   \hline
        \multirow{6}{*}{\rotatebox[origin=c]{90}{\textbf{XNLI}}}
     &   en	& 137850 & 	27	& 19269	& 0.14 \\
	&de	& 135213	& 26	&16999	& 0.13 \\
	&es	& 147127 & 	29	& 16130	& 0.11 \\
	&fr	& 152867	& 30	& 17355	& 0.11 \\
	&hi	& 159243	& 31	& 14701	& 0.09 \\
	&tr	& 104793	& 20	& 16211	& 0.15 \\
        \hline

        \multirow{5}{*}{\rotatebox[origin=c]{90}{\textbf{WikiANN}}}
       &  en	& 80326	& 8	& 8214	& 0.10 \\
	& de	& 97646	& 9	&7902	& 0.08 \\
	& es	& 64727	& 6	& 7244	& 0.11 \\
	& fr	& 68754	& 6	&  7235	& 0.11 \\
	& tr	& 75731	& 7	& 8176	& 0.11 \\
        \hline

        \multirow{6}{*}{\rotatebox[origin=c]{90}{\textbf{SNIPS}}}
        & en	& 13159	& 9	&1822	& 0.14 \\
	& de	& 13546	& 9	& 1912	& 0.14 \\
	& es	& 14411	& 10 & 1889	& 0.13 \\
	& fr	& 14323	& 10 & 1964 & 0.14  \\
	& hi	& 13968	& 9	& 1239 & 0.09 \\
	& tr	& 10329	& 7	& 1513	& 0.15 \\
        \hline
    \end{tabular}
   
    \caption{Number of tokens for each language in the test set of each dataset along with the number of noise inserted in to the test set and the noise ratio per language.
    }

    \label{tab:noise_stat}

\end{table}

\section{Experimental Setup}
In our experiments, we utilized 
four A100 and A40 GPUs to fine-tune the models. To optimize memory usage and train the models on multiple GPUs, we also employed Deepspeed\footnote{\url{https://deepspeed.ai}} alongside the Zero-3 Redundancy Optimizer \cite{rajbhandari2020zero}.

\subsection{Datasets and Tasks}
To evaluate the robustness of LLMs to real-world noisy data, we opted for 3 NLP tasks that are different from one another. Table \ref{tab:dataset_size} shows the training and test size of each dataset for each language.
\\
\noindent
\textbf{SNIPS} \cite{coucke2018snips}.
The SNIPS Natural Language Understanding dataset is a publicly available voice collection of over $16000$ user queries, categorized into seven distinct user intentions with varying degrees of complexity. This dataset is used for the Intent classification (IC).
The extracted text dataset in English is available on a HuggingFace reposetory\footnote{\url{https://huggingface.co/datasets/benayas/snips}}. Using the SeamlessM4T v2 model \cite{barrault2023seamless}, we construct a multilingual dataset of SNIPS consisting of its translation in German, Hindi, French, Spanish, and Turkish languages. 


      
    
    

\begin{table}[t]
    \centering
    \small
    \setlength{\tabcolsep}{4pt}
    \begin{tabular}{l |c| c| r}
        \hline
        \textbf{Datasets} & \textbf{Languages} & \textbf{Train}  & \textbf{Test} \\
        \hline
        WikiANN & en, de, fr, es, tr & 20000 & 10000  \\
        WikiANN & hi & 5000  & 1000  \\

        XNLI & en, de, fr, es, hi, tr & 392702  & 5010  \\
      
        SNIPS & en, de, fr, es, hi, tr & 13100 & 1400  \\
        \hline
    \end{tabular}
    
    \caption{Statistics of the utilized datasets in our experiments. The languages are English (en), German(de), French (fr), Spanish (es), Turkish (tr), and Hindi (hi). Values show the size of the train and test sets for each language in three datasets. Only the Hindi (hi) language in WikiANN dataset is smaller than the other languages.}
    \label{tab:dataset_size}
    
\end{table}

\noindent
\textbf{XNLI} \cite{conneau2018xnli}.
This dataset is used for the Natural Language Inference (NLI) task determining whether a \textit{hypothesis} is true (entailment), false (contradiction), or undetermined (neutral) given a \textit{premise}. 

\noindent
\textbf{WikiANN} \cite{rahimi2019massively}.
WikiANN (also known as PAN-X) is a multilingual named entity recognition (NER) dataset consisting of Wikipedia articles annotated with LOC (location), PER (person), and ORG (organization) tags. 


To ensure that each model is trained on the same data, we construct a combined, shuffled dataset containing all six languages (five languages for WikiANN) in our study. 

\subsection{Models}
To understand how the complexity of a model (measured by the number of parameters) impacts its performance against noisy input, we opted for 9 models of various sizes, from mBERT-base with close to 0.2B parameters to the largest version of the mT5 model with 13B parameters. 

\noindent
\textbf{mBERT-base} \cite{devlin2018bert}.
This model has been trained on the top 100 largest languages in Wikipedia and has a Transformer-based (encoder-only) structure.

\noindent
\textbf{XLM-RoBERTa-base} \cite{conneau2019unsupervised}.
This is a Transformer-based multilingual masked language model trained on 2.5TB of filtered CommonCrawl data.

\begin{figure}[t]
\centering
\includegraphics[scale=0.35]{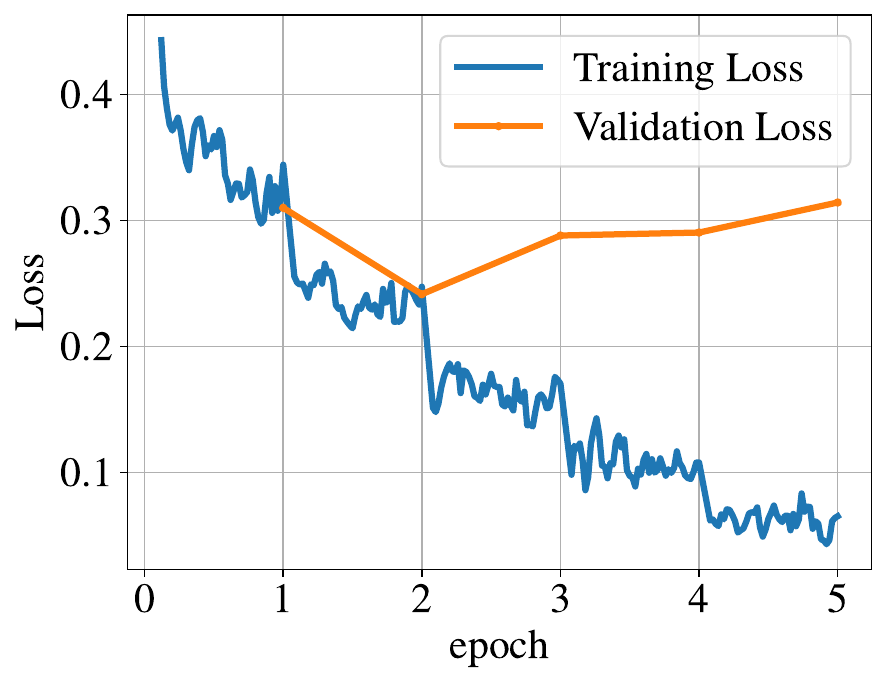}
\caption{Training and evaluation losses for NER task on WikiANN dataset. After the second epoch model overfits. 
}
\label{fig:losses}

\end{figure}

\noindent
\textbf{mT5} \cite{xue2020mt5}.
This model's design is similar to the T5 model \cite{raffel2020exploring}, which uses a basic encoder-decoder Transformer architecture \cite{vaswani2017attention} trained on mC4 corpus\footnote{\url{https://huggingface.co/datasets/allenai/c4}}.
We evaluate five versions of the mT5 model listed in Table \ref{tab:models_list} in the appendix.

\noindent
\textbf{Falcon} \cite{almazrouei2023falcon}.
The architecture of the Falcon models is based on PaLM which uses a standard Transformer model architecture in a decoder-only setup \cite{chowdhery2023palm} and has been trained on RefinedWeb dataset \cite{penedo2023refinedweb}. This model mainly supports English, German, French, and Spanish languages.
We use the Falcon-7B model in our benchmarks.

\noindent
\textbf{BLOOM} \cite{le2022bloom}.
The BLOOM model architecture is based on the causal-decoder Transformer model trained on multilingual data with 350 billion tokens data. Although 46 languages were supported, BLOOM lacks compatibility with German and Turkish languages. We use the 7B version of the BLOOM model.


\subsection{Fine-tuning}


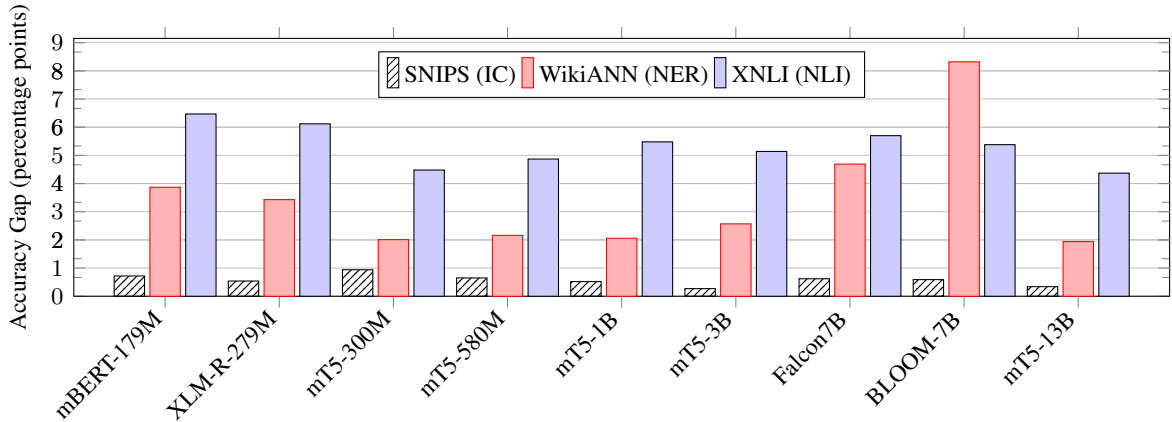
\begin{figure*}[t]
\centering
\begin{tikzpicture}

\begin{axis}[
    ybar,
    width=\textwidth,
    height=5cm,
    ymin=0,
    ylabel={Accuracy Gap (percentage points)},
    y label style={font=\small},
    y tick label style={font=\small},
    xlabel={ },
    symbolic x coords={mBERT-179M, XLM-R-279M, mT5-300M, mT5-580M, mT5-1B, mT5-3B, Falcon7B, BLOOM-7B, mT5-13B},
    xtick=data,
    legend style={at={(0.5,0.95)},
      anchor=north,legend columns=-1},
    enlarge x limits=0.1,
    bar width=0.4cm,
    x tick label style={rotate=45, anchor=east, font=\small}, 
    ymajorgrids=true,
    grid style={line width=.1pt, draw=gray!10},
    major grid style={line width=.2pt,draw=gray!50},
    minor tick num=2,
    legend image code/.code={
        \draw [#1] (0cm,-0.15cm) rectangle (0.15cm,0.25cm); },
    legend style={anchor=north,legend columns=-1,font=\small},
    extra y ticks={1,2,3,4,5,6,7,8,9},  
    extra y tick style={grid=major},
    ]
\addplot [pattern=north east lines] coordinates {(mBERT-179M, 0.72) (XLM-R-279M, 0.54) (mT5-300M, 0.94) (mT5-580M, 0.65) (mT5-1B, 0.52) (mT5-3B, 0.27) (mT5-13B, 0.34) (Falcon7B, 0.62) (BLOOM-7B, 0.59) };

\addplot coordinates {(mBERT-179M, 3.87) (XLM-R-279M, 3.43) (mT5-300M, 2.01) (mT5-580M, 2.16) (mT5-1B, 2.06) (mT5-3B, 2.57) (mT5-13B, 1.94) (Falcon7B, 4.69) (BLOOM-7B, 8.32)};

\addplot [fill=blue!20]
     coordinates {(mBERT-179M, 6.47) (XLM-R-279M, 6.12) (mT5-300M, 4.48) (mT5-580M, 4.87) (mT5-1B, 5.48) (mT5-3B, 5.14) (mT5-13B, 4.37) (Falcon7B, 5.70) (BLOOM-7B, 5.38)};
\legend{SNIPS (IC), WikiANN (NER), XNLI (NLI)}
\end{axis}
\end{tikzpicture}
\caption{Average gap (in percentage points) between the accuracy of the experimented models on the clean data and the noisy data. The numbers indicate the average gap over all the six languages on SNIPS, Wikiann, and XNLI datasets.}
\label{fig:c-n-average}
\end{figure*}

To prevent overfitting, we have examined the validation and training loss values for different languages and models, revealing that 2 epochs are enough in most scenarios. Figure \ref{fig:losses} depicts the training and evaluation losses for the WikiANN dataset running for six epochs. For some models that required more training to show convergence, we performed fine-tuning for six epochs. Table \ref{tab:hyperparams} in the appendix contains detailed information about hyperparameters. We used the same configuration when doing the experiments on the clean and noisy sets to perform a fair comparison of these data. 

\section{Results and Findings}\label{sec:results}

We fine-tune the models on WikiANN (NER) and SNIPS (IC) three times and on the XNLI (NLI) two times (due to the size of this dataset which is computationally expensive) and choose the best model based on its clean performance for evaluating on the test sets. We investigate the performance gap of the clean and noisy test sets (C-N) for models and languages and the averaged values.




\begin{table*}[t]
\centering
\small
\begin{tabularx}{\textwidth}{|l|X|X|X|X|X|X|X|X|}
\hline
\multirow{2}{*}{Models} & \multirow{2}{*}{} & \multicolumn{6}{c|}{XNLI (NLI) - Accuracy(\%)} & \multirow{2}{*}{Average} \\
\cline{3-8}
& & en & de & es & fr & hi & tr & \\
\hline
\multirow{3}{*}{mBERT-179M} & Clean & 82.55 & 78.24 & 79.56 & 78.82 & 69.90 & 73.37 & 77.07 \\
& Noisy & 71.94 & 72.51 & 71.50 & 72.10 & 66.93 & 68.64 & 70.60 \\
 &  C-N & 10.61 & 5.73 & 8.06 & 6.72 & 2.97 & 4.73 & 6.47 \\
\hline
\multirow{3}{*}{XLM-R-279M} & Clean & 84.79 & 80.06 & 81.94 & 80.68 & 74.35 & 77.29 & 79.85 \\
& Noisy & 75.21 & 74.21 & 75.23 & 75.03 & 70.86 & 71.86 & 73.73 \\
&  C-N & 9.58 & 5.85 & 6.71 & 5.65 & 3.49 & 5.43 & 6.12 \\
\hline
\multirow{3}{*}{mT5-300M} & Clean & 73.91 & 68.76 & 71.12 & 70.02 & 63.83 & 66.81 & 69.08 \\
& Noisy & 65.83 & 65.87 & 66.09 & 65.01 & 61.92 & 62.85 & 64.60 \\
&  C-N & 8.08 & 2.89 & 5.03 & 5.01 & 1.91 & 3.96 & 4.48 \\
\hline
\multirow{3}{*}{mT5-580M} & Clean & 84.45 & 79.82 & 81.36 & 80.38 & 74.67 & 76.25 & 79.49 \\
& Noisy & 74.25 & 74.71 & 75.07 & 75.55 & 69.56 & 72.55 & 73.61 \\
&  C-N &10.20 & 5.11 & 6.29 & 4.83 & 5.11 & 3.70 & 5.87 \\
\hline
\multirow{3}{*}{mT5-1B} & Clean & 88.82 & 83.97 & 85.03 & 84.33 & 79.40 & 81.26 & 83.80 \\
& Noisy & 80.38 & 78.88 & 79.64 & 79.52 & 74.27 & 77.23 & 78.320 \\
&  C-N & 8.44 & 5.09 & 5.39 & 4.81 & 5.13 & 4.03 & 5.48 \\
\hline
\multirow{3}{*}{mT5-3B} & Clean & 90.1 & 86.47 & 87.03 & 86.91 & 81.82 & 83.91 & 86.04 \\
& Noisy & 83.09 & 81.36 & 82.0 & 81.92 & 77.66 & 79.4 & 80.90 \\
&  C-N & 7.01 & 5.11 & 5.03 & 4.99 & 4.16 & 4.51 & 5.14 \\
\hline

\multirow{3}{*}{Falcon7B} & Clean & 90.68 & 84.63 & 86.31 & 86.25 & - & - & 86.97 \\
& Noisy & 84.59 & 78.54 & 80.42 & 81.54 & - & - & 81.27 \\
&  C-N & 6.09 & 6.09 & 5.89 & 4.71 & - & - & 5.70 \\
\hline
\multirow{3}{*}{BLOOM-7B} & Clean & 89.52 & - & 86.63 & 85.73 & 78.32 & - & 85.05 \\
& Noisy & 82.46 & - & 81.44 & 80.98 & 73.81 & - & 79.67 \\
&  C-N & 7.06 & - & 5.19 & 4.75 & 4.51 & - & 5.38 \\
\hline
\multirow{3}{*}{mT5-13B} & Clean & 91.28 & 87.23 & 88.12 & 87.29 & 84.33 & 85.25 & 87.25 \\
& Noisy & 86.03 & 82.87 & 83.91 & 83.43 & 79.22 & 81.84&  82.88 \\
&  C-N & 5.25 & 4.36 & 4.21 & 3.86 & 5.11 & 3.41 & 4.37 \\

\hline

\end{tabularx}

\caption{Performance (F1 score) of each model on the five languages. \texttt{Clean} means clean data, \texttt{Noisy} means our created noisy data, and \texttt{C-N} means the difference between the performance on the clean and noisy data. English exhibits the highest performance gap among the languages, due to the quantity and types (PoS) of noise inserted.}
\label{tab:xnli_results}
\end{table*}


\subsection{Are larger models more robust?}\label{sub:models}
To analyze the overall performance of the selected models, we average the results (performance gap) over all the languages and datasets. Looking at the mT5 models which have the same components with only difference in size, we see that the largest model (mT5-13B) is also the most robust while the smallest model (mT5-300M) is the most vulnerable. 
Considering different models with various architectures and tokenizers, Tables \ref{tab:average_gap_mT5} and \ref{tab:average_gap_other} show that mT5 models are less vulnerable to typos than other models with only 2.27 percent degradation in performance, while Falcon and BLOOM perform poorly despite having larger sizes, with 3.67 and 4.27 percent degradation, respectively. This can be attributed to two reasons. Looking at Table \ref{tab:Training_languages_orig}, we hypothesize that the main reason is the amount of training data used for each of these models. The mT5 model has by far seen the largest number of tokens in all the languages. This could have helped the model to be more resistant to small perturbations in the input. However, Table \ref{tab:Training_languages_orig} also shows that Falcon and BLOOM models have seen more data than BERT-like models but perform worse than them. This leads us to the second reason which is their language modeling paradigm. Given their decoder-only structure, they can have difficulty with some tasks such as named entity recognition compared to mBERT and XLM-R which use masked language modeling (see \S\ref{fig:c-n-average}). 

\begin{mytheo}{}{}
  mT5 models show more robustness and the largest mT5 model is the most robust.
\end{mytheo}

\begin{table}[t]
\centering
\small
\setlength{\tabcolsep}{2pt}
\begin{tabular}{l r}

    \hline

    \textbf{Model} & \textbf{Average gap} \\
           
    \hline
    mT5-13B & \texttt{2.27} \\
    mT5-300M &2.47 \\
    mT5-3B & 2.64 \\
    mT5-1B &   2.76 \\
    mT5-580M & 2.95 \\
    \hline
\end{tabular}
\caption{Average performance gap over all the languages and all the datasets in increasing order for mT5 models }
\label{tab:average_gap_mT5}
\end{table}
\begin{table}[t]
\centering
\small
\setlength{\tabcolsep}{2pt}
\begin{tabular}{l r}

    \hline

    \textbf{Model} & \textbf{Average gap} \\
           
    \hline

    XLM-R-279M & 3.29 \\
    mBERT-179M & 3.58 \\
    Falcon-7B & 3.67 \\
    BLOOM-7B & \textbf{4.27} \\
    
    \hline
\end{tabular}
\caption{Average performance gap over all the languages and all the datasets in increasing order for XLMR, mBERT, Falcon, BLOOM models.}
\label{tab:average_gap_other}
\end{table}

\begin{table}[t]
    \centering
    \small
    \setlength{\tabcolsep}{2.75pt}
    
    \begin{tabular}{l | c| c| c| c| c}
    \hline
   \textbf{Languages} &  \textbf{mT5} & \textbf{XLM-R} & \textbf{Falcon} & \textbf{mBERT} & \textbf{BLOOM}\\
   \hline

English & 2733 & 0.30 & 750 &  4.59 & 112 \\
Spanish  & 433 & 0.05 & 17 & 1.20 & 40 \\
German  & 347 & 0.07 & 18 & 1.56 & - \\
French  & 318 & 0.06 & 16 & 1.68 & 46 \\
Turkish  & 71 & 0.02 & - & 0.16 & - \\
Hindi  & 24 & 0.02 & - & 0.06 & 5.6 \\
        \hline
    \end{tabular}
   
    \caption[Caption for LOF]{
    Number of tokens used in the training dataset of each model for each language. The data for mT5, XLM-R (XLM-RoBERTA), and the BLOOM models were collected from their main article.}

    \label{tab:Training_languages_orig}

\end{table}

\subsection{Are different tasks equally sensitive to real-world noise?}\label{sub:tasks}

\begin{figure}[t]
\includegraphics[scale=0.33]
{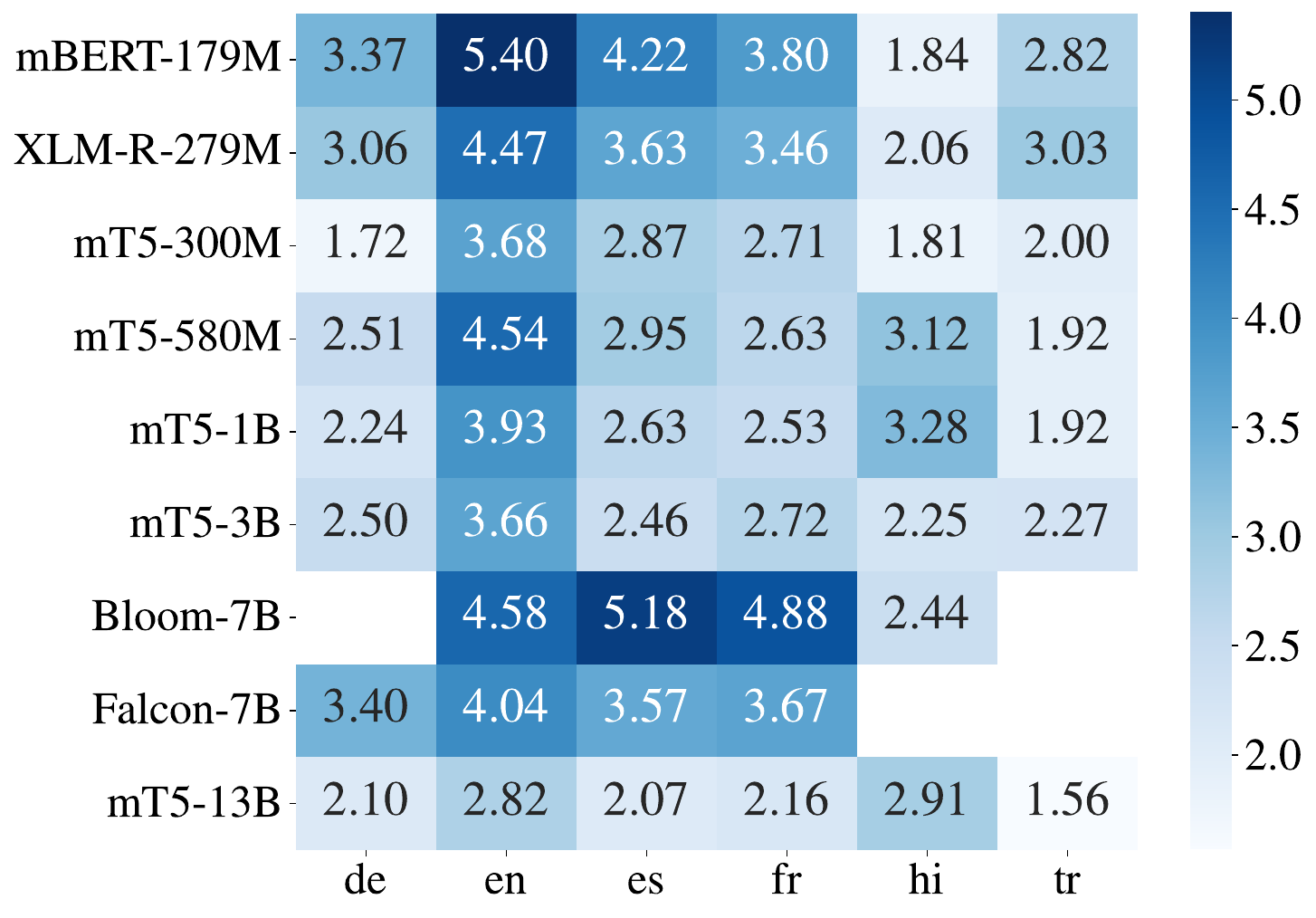}
\caption{Heatmap of average performance gap over datasets models per language.}
\label{fig:language_corr}
\end{figure}

Figure \ref{fig:c-n-average} compares the average performance gap over the languages for the individual datasets. For the figure, we can see that, at around 5 percentage points ($\pm 1$), the performance decrease is similar for all the models in the NLI task as well as the intent classification (IC) task, where we see less than one percent degradation. Looking at Table \ref{tab:noise_stat}, we can justify the similarly big gap in the NLI task across all models. Since the samples are longer in the XNLI dataset, there are more noisy words inserted in each sample (around 3 times the noise in other datasets). As a result, all models show similar vulnerability in this task. However, in the NER task, the performance drop greatly varies across the models from around 2 percent for mT5 models to more than 8 percent for the BLOOM model and a mid-range decrease for the BERT-like models. It is also worth mentioning that the second largest gap belongs to the other decoder-only model in our experiments, Falcon, with close to a 5 percent performance drop, making the two decoder-only models, the least robust in the NER task. The reason for such discrepancy can be attributed to the training algorithm for these models, being next-token prediction, compared to the algorithm for other models which is masked language modeling (MLM). It seems that it is easier to perform the NER task with the MLM paradigm than the next-token prediction. 

\begin{mytheo}{}{}
  NLI task is more sensitive to real-world noise (WikiTypo).
\end{mytheo}

\subsection{How does the performance differ from English to other languages?}

To look at the performance gap for each language, we average the results over the datasets. Figure \ref{fig:language_corr} shows the heatmap of these results, where we can see that almost all models perform poorly on English than on other languages. A similar pattern was also observed by \cite{stickland2023robustification}, where the authors attributed this to the higher performance of these models on English compared to others. Indeed, this can be confirmed by looking at the absolute performance numbers in Table \ref{tab:xnli_results}. Further results for other datasets can be seen in Tables \ref{snips_results} and \ref{tab:wikiann_results} in the Appendix.

\begin{mytheo}{}{}
  English is less robust to WikiTypo than other languages.
\end{mytheo}

\subsection{Noisy training narrows the gap}

Introducing noise to the training data can boost the robustness of a language model \cite{pan2024can}. To show this for our setting, we fine-tune the BLOOM model on the WikiANN which has shown the largest performance drop. Interestingly, as shown in Table \ref{tab:noisy_training}, fine-tuning the BLOOM model on only the noisy training data reduces the performance gap on the languages in the WikiANN dataset. However, while the reduction is partly due to the increase in the performance of the model on the noisy test data, it is also partly because of the performance drop on the clean data. 

\begin{table}[t]
    \centering
    \small
    \setlength{\tabcolsep}{3pt}
    \begin{tabular}{l |c| c c c}
        \hline
        & &\textbf{en} & \textbf{es} & \textbf{fr}\\
        \hline
        \multirow{3}{*}{Clean Train} 
       & Clean & 45.68 & 66.39 & 62.23  \\
       & Noisy & 39.56 & 56.75 & 53.05  \\
       & C-N &6.12 &	9.64	& 9.18       \\[3pt]
       \hline
        \multirow{3}{*}{Noisy Train} 
       & Clean &  43.45 & 64.41 & 60.01 \\
       & Noisy & 40.09 & 59.49 & 54.68\\
       & C-N &  3.36 &	4.92&	5.33 \\[3pt]
       \hline 
    \textbf{Improvement} & C-N & 2.76&	4.72&	3.85\\

        \hline
    \end{tabular}
    
    \caption{Bloom-7B model's clean and noisy performance and the gap between them after evaluating two sets of experiments. 1) Fine-tuning the model with the original training set. 2) Fine-tuning the model after inserting noise into the training set. The gap is decreased after fine-tuning the model with the noisy training data.}
    \label{tab:noisy_training}
    
\end{table}

\subsection{Why are models less robust to WikiTypo English noise?}

To investigate the performance gap of the different languages, we look into the noise in the grammatical categories introduced into our largest dataset (XNLI). We analyzed the distribution of part-of-speech (PoS) tags within the injected noise in the test set, using spaCy\footnote{\url{https://spacy.io}}. Table \ref{tab:pos_xnli} presents the POS tag distribution across the four languages supported by spaCy.
Our analysis revealed that there are almost twice as many noisy verb instances in English compared to other languages.  Furthermore, the number of noisy nouns and proper nouns is also higher in English. Earlier studies have demonstrated that replacing the adverbs and adjectives has less impact on the semantics of the sentence than the verbs and nouns \cite{periti2024analyzing}. This observation aligns with the English language's higher performance gap on the XNLI dataset. 

Another compelling finding is the disproportionate impact of noise on the English language compared to other languages, particularly in smaller model sizes. On the XNLI dataset, the performance gap between clean and noisy data is significantly larger for English models compared to their counterparts (see Table \ref{tab:xnli_results}). However, as model size increases, this disparity tends to diminish, with English performance converging towards the average performance across all languages.

\begin{table}[t]
    \centering
    \small
    \setlength{\tabcolsep}{4pt}
    \begin{tabular}{l |c | c | c| c}
        \hline
        \textbf{PoS} & \textbf{English} & \textbf{Spanish}  & \textbf{German} & \textbf{French}\\
        \hline
        \rowcolor{lightgray} ADJ & 1117 & 1665 & 790 & 1948 \\
        ADP & 2102 & 2112 & 2323 & 1918 \\
         \rowcolor{lightgray} ADV & 1233 & 880 & 1776 & 1662 \\
        AUX & 1707 & 1183 & 1570 & 721 \\
        CCONJ & 164 & 61 & 524 & 492 \\
        DET & 901 & 2513 & 2200 & 2429 \\
        INTJ & 216 & 35 & 0 & 14 \\
        \rowcolor{lightgray} NOUN & 4767 & 3302 & 1863 & 3927 \\
        NUM & 155 & 131 & 39 & 72 \\
        PART & 1646 & 2 & 269 & 0 \\
        PRON & 1032 & 1154 & 2133 & 1190 \\
        PROPN & 799 & 119 & 522 & 344 \\
        PUNCT & 6 & 2 & 0 & 0 \\
        SCONJ & 161 & 729 & 284 & 387 \\
        SYM & 0 & 2 & 0 & 0 \\
       \rowcolor{lightgray} VERB & 4475 & 2240 & 2537 & 2227 \\
        Other & 11 & 0 & 169 & 24 \\
        \hline
    \end{tabular}
    \caption{ Distribution of part of speech for all the noises inserted in the XNLI noisy test set. Number of noisy verbs injected to the test sets are almost twice as much in English than other languages.}
    \label{tab:pos_xnli}
\end{table}

\section{Conclusion}
We build a corpus of real-world noise (typos) from Wikipedia edits history and explore how the noisy data impacts the performance of LLMs across languages. Performance gap across six languages indicates that all models exhibit vulnerability when encountering noisy input. In addition, our findings suggest that the robustness of language models against noisy input is influenced by various factors, including the size and language coverage of their training data, their underlying architectural design and parameter count, and the specific task they are evaluated on. While mT5 models demonstrate the best performance against noise likely due to their massive training data, size was not the sole factor. Architecture-wise, decoder-only models such as BLOOM and Falcon show more vulnerability to noise, especially in the NER task. These findings highlight the importance of considering all model and dataset features in evaluating the robustness of language models.

\section{Limitations}

While we evaluated a broad range of models (200M to 13B parameters), current capabilities extend beyond this. Due to time and computational constraints, we couldn't explore even larger models. Our work focused on six languages. However, recent multilingual models can handle over 100 languages. Therefore, a complete assessment of LLM robustness should ideally consider more languages. We utilized typos from Wikipedia edits as our realistic noise source. While effective, exploring other noise sources and data-gathering methods for real-world noise could further strengthen the research. We presented results using a noise insertion ratio of 0.2. To gain a more comprehensive understanding of model robustness to noise, various noise levels could be explored. Our study investigated three NLP tasks, but a broader exploration of different tasks using diverse datasets would depict a greater perspective on the model's reliability. 
Finally, While the WikiTypo noise collection contains over 40,000 typos, it is limited in proper noun examples due to the relatively infrequent occurrence of such errors in Wikipedia articles. Therefore, due to the high concentration of proper nouns in the WikiANN dataset, we employed different strategies for generating the noisy test set.


\bibliography{custom}
\newpage
\appendix

\section{Models' Characteristics}\label{sec:appendix}
Tables \ref{tab:tokenizer} and \ref{tab:models_list} show each model's tokenizer algorithm and the number of parameters and table \ref{tab:Training_languages} show the proportion of each language used in training the models.
 \begin{table}[h]
    \centering
    \small
    \setlength{\tabcolsep}{2pt}
    \begin{tabular}{l r}
        \hline
        \textbf{Model} & \textbf{Tokenizer Techniqe}\\
        \hline
        mBERT &  WordPiece \\
        XLM-RoBERTa &  SentencePiece \\

        mT5 &   SentencePiece\\
        Falcon & Byte-level BPE\\
        BLOOM & Byte-level BPE  \\

        \hline
    \end{tabular}
    
    \caption{Tokenizer algorithm for the models evaluated in this study.
    BPE stands for Byte Pair Encoding. }
    \label{tab:tokenizer}
    
\end{table}

    
    

\begin{table}[h]
    \centering
    \small
    \setlength{\tabcolsep}{5pt}
    \begin{tabular}{l r}
        \hline
        \textbf{Model} & \textbf{\# Params}\\
        \hline
        mBERT-base &  179M \\
        XLM-RoBERTa-Base &  279M  \\
        mT5-small & 300M   \\
        mT5-base & 580M   \\
        mT5-large & 1.2B   \\
        mT5-xl & 3.7B   \\
        
        Falcon-7b & 7B   \\
        BLOOM-7b & 7B   \\
        mT5-xxl & 13B   \\

        \hline
    \end{tabular}
    
    \caption{Studied models' size based on the number of parameters in million (M) and billion (B).}
    \label{tab:models_list}
    
\end{table}

\begin{table}[h]
    \centering
    \small
    \setlength{\tabcolsep}{3.5pt}
    
    \begin{tabularx}{\columnwidth}{l |c| c| c| c| c}
    
        \hline
\textbf{Language} &  \textbf{mT5} & \textbf{XLMR} & \textbf{Falcon} & \textbf{mBERT} & \textbf{BLOOM}\\
\hline
English & 5.67 & 12.14 & 46.43 & 10.84 & 33.68 \\
Spanish  & 3.09 & 2.15 & 4.62 & 3.10 & 12.16 \\
German  & 3.05 & 2.69 & 5.84 & 4.62 & - \\
French  & 2.89 & 2.29 & 4.64 & 4.15 & 14.46 \\
Turkish  & 1.93 & 0.84 & - & 0.96 & - \\
Hindi  & 1.21 & 0.81 & - & 0.26 & 1.71 \\
        \hline
    \end{tabularx}
    
    \caption{Percent of languages (in our study) used in the training dataset of each model. English has the highest ratio among the languages in the pre-training process. }
    \label{tab:Training_languages}
    
\end{table}
\section{Results} \label{sec:appendix}
The complete results of the SNIPS and WikiANN datasets are available in Tables \ref{snips_results}, \ref{tab:wikiann_results}. Figure \ref{all_languages_performance} indicates the clean and noisy performance and the gap between them for each of the six languages.

\begin{figure*}
    \centering
    \begin{subfigure}{.5\textwidth}
        \centering
        \includegraphics[width=\linewidth]{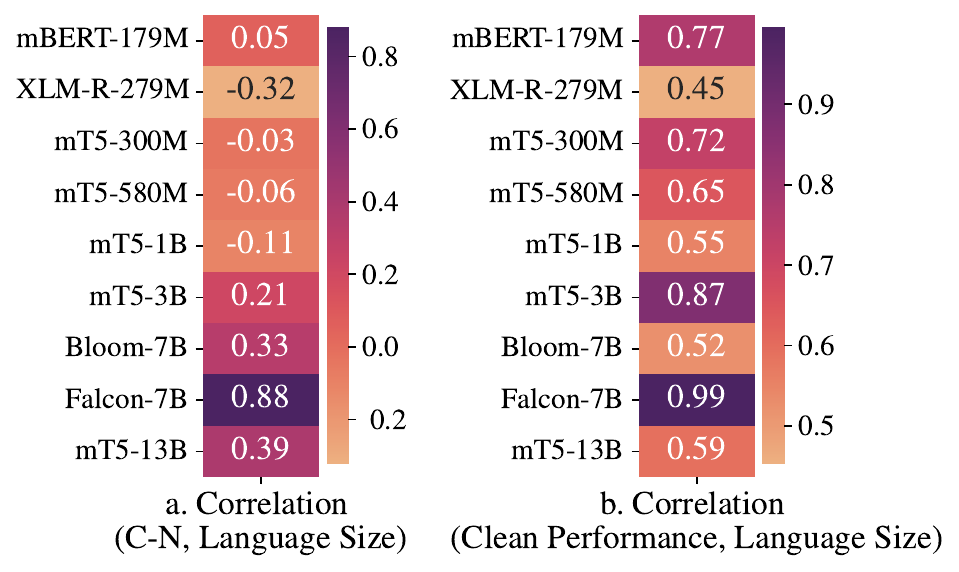}
        \caption{SNIPS (IC)}
    \end{subfigure}%
    \begin{subfigure}{.5\textwidth}
        \centering
        \includegraphics[width=\linewidth]{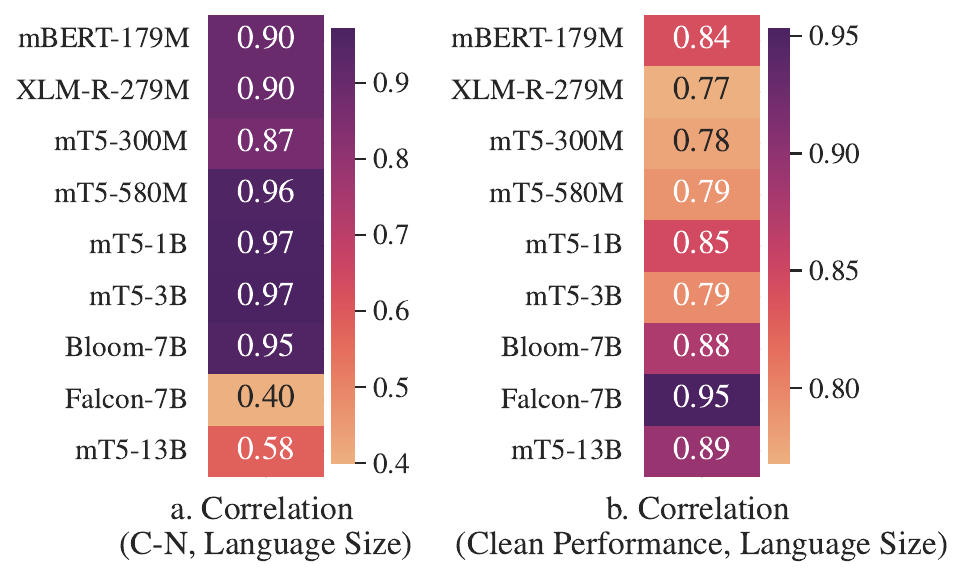}
        \caption{XNLI (NLI)}
    \end{subfigure}
    \begin{subfigure}{.5\textwidth}
        \centering
        \includegraphics[width=\linewidth]{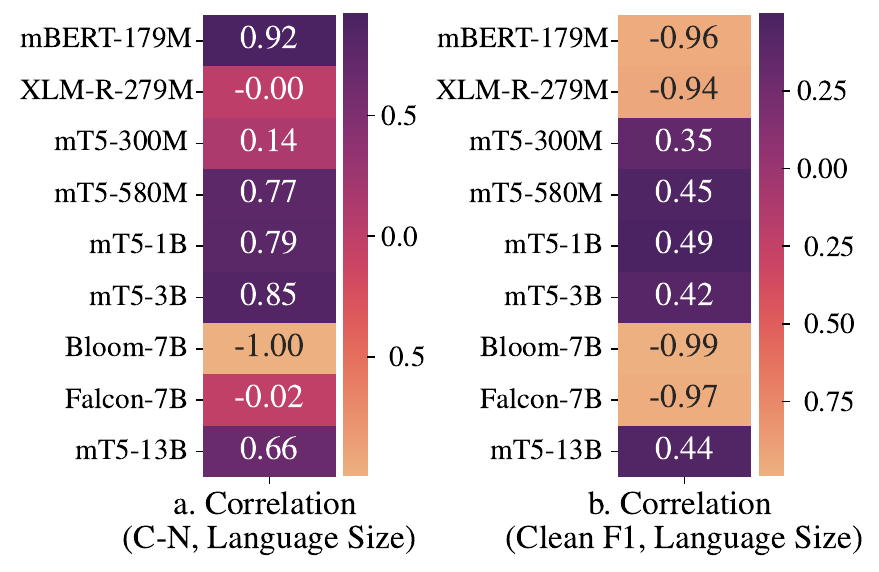}
        \caption{WikiANN (NER)}
    \end{subfigure}
    \caption{The heatmaps show the (Pearson) correlation of the performance gap and the model's clean performance with the language size of the models over three SNIPS, XNLI, and WikiANN datasets. The correlation varies across different datasets (tasks). However, clean performance generally exhibits a positive correlation with language model size, except for the WikiANN dataset when considering models other than mT5.}
    \label{fig:corr_heatmaps}
\end{figure*}

Figure \ref{fig:corr_heatmaps} shows the impact of the amount of each language used in training on the clean performance and the gap between clean and noisy for the SNIPS, WikiANN, and XNLI datasets.
The experiments fine-tuning details and hyperparameters are shown in Table \ref{tab:hyperparams}.
\begin{table*}[h]

\centering
\small
\begin{tabularx}{\textwidth}{|l|X|X|X|X|X|X|X|X|}
\hline
\multirow{2}{*}{Models} & \multirow{2}{*}{language} & \multicolumn{6}{c|}{SNIPS (IC) - Accuracy(\%)} & \multirow{2}{*}{Average} \\
\cline{3-8}
& & en & de & es & fr & hi & tr& \\
\hline
\multirow{3}{*}{mBERT-179M} & Clean & 99.07 & 99.07 & 98.71 & 98.64 & 98.57 & 98.36 & 98.74 \\
& Noisy & 98.29 & 98.29 & 97.93 & 98.14 & 97.86 & 97.57 & 98.01 \\

&   C-N & 0.78 & 0.78 & 0.78 & 0.5 & 0.71 & 0.79 & 0.72 \\
\hline
\multirow{3}{*}{XLM-R-279M} & Clean & 99.0 & 98.93 & 99.0 & 99.07 & 98.71 & 98.57 & 98.88 \\
& Noisy & 98.57 & 98.71 & 98.29 & 98.36 & 98.07 & 98.07 & 98.34 \\
&   C-N & 0.43 & 0.22 & 0.71 & 0.71 & 0.64 & 0.5 & 0.54 \\
\hline
\multirow{3}{*}{mT5-300M} & Clean & 98.71 & 98.36 & 97.93 & 97.86 & 97.0 & 96.86 & 97.79 \\
& Noisy &97.79 & 97.71 & 96.64 & 97.07 & 95.29 & 96.57 & 96.84 \\
&   C-N &0.92 & 0.65 & 1.29 & 0.79 & 1.71 & 0.29 & 0.94 \\
\hline
\multirow{3}{*}{mT5-580M} & Clean &99.07 & 98.71 & 98.86 & 98.64 & 97.71 & 97.86 & 98.48 \\
& Noisy & 98.43 & 98.29 & 98.07 & 98.07 & 96.57 & 97.5 & 97.82 \\
&   C-N & 0.64 & 0.42 & 0.79 & 0.57 & 1.14 & 0.36 & 0.65 \\
\hline
\multirow{3}{*}{mT5-1B} & Clean & 98.79 & 98.57 & 98.07 & 98.43 & 98.64 & 98.29 & 98.46 \\
& Noisy & 98.29 & 98.43 & 97.57 & 98.07 & 97.21 & 98.07 & 97.94 \\
&   C-N & 0.5 & 0.14 & 0.5 & 0.36 & 1.43 & 0.22 & 0.52 \\
\hline
\multirow{3}{*}{mT5-3B} & Clean & 99.29 & 99.0 & 98.71 & 98.79 & 98.71 & 98.86 & 98.89 \\
& Noisy & 98.93 & 98.5 & 98.71 & 98.5 & 98.36 & 98.71 & 98.62 \\
&   C-N & 0.36 & 0.5 & 0.0 & 0.29 & 0.35 & 0.15 & 0.27 \\
\hline

\multirow{3}{*}{Falcon-7B} & Clean & 98.93 & 97.71 & 97.64 & 97.79 & - & - & 98.02 \\
& Noisy & 97.57 & 97.36 & 97.57 & 97.07 & - & - & 97.39 \\
&   C-N & 1.36 & 0.35 & 0.07 & 0.72 & - & - & 0.62 \\
\hline
\multirow{3}{*}{BLOOM-7B} & Clean &98.5 & - & 98.71 & 98.57 & 97.86	&-  & 98.41 \\
& Noisy & 97.93 & - & 98.0 & 97.86 & 97.5&- & 97.82 \\
&   C-N & 0.57 & - & 0.71 & 0.71 & 0.36 & -& 0.59 \\

\hline

\multirow{3}{*}{mT5-13B} & Clean & 99.29 & 99.21 & 99.14 & 99.0 & 98.5 & 98.93 & 99.01 \\
& Noisy & 98.71 & 98.86 & 98.86 & 98.93 & 97.79 & 98.86 & 98.67 \\
&   C-N & 0.58 & 0.35 & 0.28 & 0.07 & 0.71 & 0.07 & 0.34 \\

\hline

\end{tabularx}

    \caption{SNIPS dataset clean and noisy test results form the WikiTypo noise insertion method for each model and each language. C-N shows the performance degradation amount. Hindi exhibits the highest performance gap, particularly in mT5 models, while Turkish demonstrates the least. }
    \label{snips_results}
\end{table*}

\begin{table*}[t]
\centering
\small
\begin{tabularx}{\textwidth}{|l|X|X|X|X|X|X|X|}

\hline

\multirow{2}{*}{Models} & \multirow{2}{*}{} & \multicolumn{5}{c|}{WikiANN (NER) - (F1)} & \multirow{2}{*}{Average} \\

\cline{3-7}
& & en & de & es & fr & tr & \\

\hline

\multirow{3}{*}{mBERT-179M} & Clean & 84.67 & 89.73 & 92.4 & 91.37 & 92.66 & 90.17\\
& Noisy & 79.87 & 86.12 & 88.57 & 87.19 & 89.72 & 86.29 \\
 &  C-N & 4.8 & 3.61 & 3.83 & 4.18 & 2.94 & 3.87\\

\hline

\multirow{3}{*}{XLM-R-279M} & Clean & 82.58 & 87.14 & 90.61 & 89.27 & 91.55 & 88.23 \\
& Noisy & 79.18 & 84.03 & 87.14 & 85.24 & 88.4 & 84.80\\
&  C-N & 3.4 & 3.11 & 3.47 & 4.03 & 3.15 & 3.43 \\

\hline

\multirow{3}{*}{mT5-300M} & Clean & 42.46 & 28.62 & 37.76 & 47.09 & 34.55 & 38.10 \\
& Noisy & 40.41 & 26.99 & 35.47 & 44.76 & 32.81 & 36.09\\
&  C-N & 2.05 & 1.63 & 2.29 & 2.33 & 1.74 & 2.01\\

\hline

\multirow{3}{*}{mT5-580M} & Clean & 54.45 & 39.08 & 44.14 & 56.36 & 47.44 & 48.29	 \\
& Noisy & 51.66 & 37.07 & 42.37 & 53.86 & 45.73 & 46.14\\
&  C-N & 2.79 & 2.01 & 1.77 & 2.5 & 1.71 & 2.16\\

\hline

\multirow{3}{*}{mT5-1B} & Clean & 59.54 & 44.15 & 47.03 & 60.86 & 51.47 & 52.61\\		
& Noisy & 56.7 & 42.65 & 45.02 & 58.45 & 49.95 & 50.55\\
& C-N & 2.84 & 1.5 & 2.01 & 2.41 & 1.52 & 2.06\\

\hline

\multirow{3}{*}{mT5-3B} & Clean & 54.97 & 38.29 & 45.04 & 57.28 & 48.38 & 48.79\\
& Noisy & 51.37 & 36.41 & 42.7 & 54.4 & 46.22 & 46.22\\
&  C-N & 3.6 & 1.88 & 2.34 & 2.88 & 2.16 & 2.57\\

\hline

\multirow{3}{*}{Falcon-7B} & Clean & 42.25 & 52.04 & 55.25 & 53.15 & - &  50.67\\
& Noisy & 37.58 & 48.27 & 50.49 & 47.57 & - & 45.98 \\
&  C-N & 4.67 & 3.77 & 4.76 & 5.58 & - & 4.69 \\

\hline

\multirow{3}{*}{BLOOM-7B} & Clean & 45.69 & - & 66.40 & 62.24 & - & 58.11\\
& Noisy & 39.57 & - & 56.76 & 53.05 & - & 49.79 \\
&  C-N & 6.12 & - & 9.64 & 9.19 & - & 8.32 \\

\hline

\multirow{3}{*}{mT5-13B} & Clean & 52.63 & 32.4 & 42.11 & 54.81 & 44.07 & 45.20 \\
& Noisy & 50 & 30.81 & 40.4 & 52.25 & 42.86 & 43.26 \\
&  C-N & 2.63 & 1.59 & 1.71 & 2.56 & 1.21 & 1.94 \\

\hline

\end{tabularx}
\caption{Performance (F1 score) of each model on the five languages. \texttt{Clean} means clean data, \texttt{Noisy} means our created noisy data, and \texttt{C-N} means the difference between the performance on the clean and noisy data. Decoder-only models (BLOOM and Falcon) are more vulnerable to noise}
\label{tab:wikiann_results}
\end{table*}
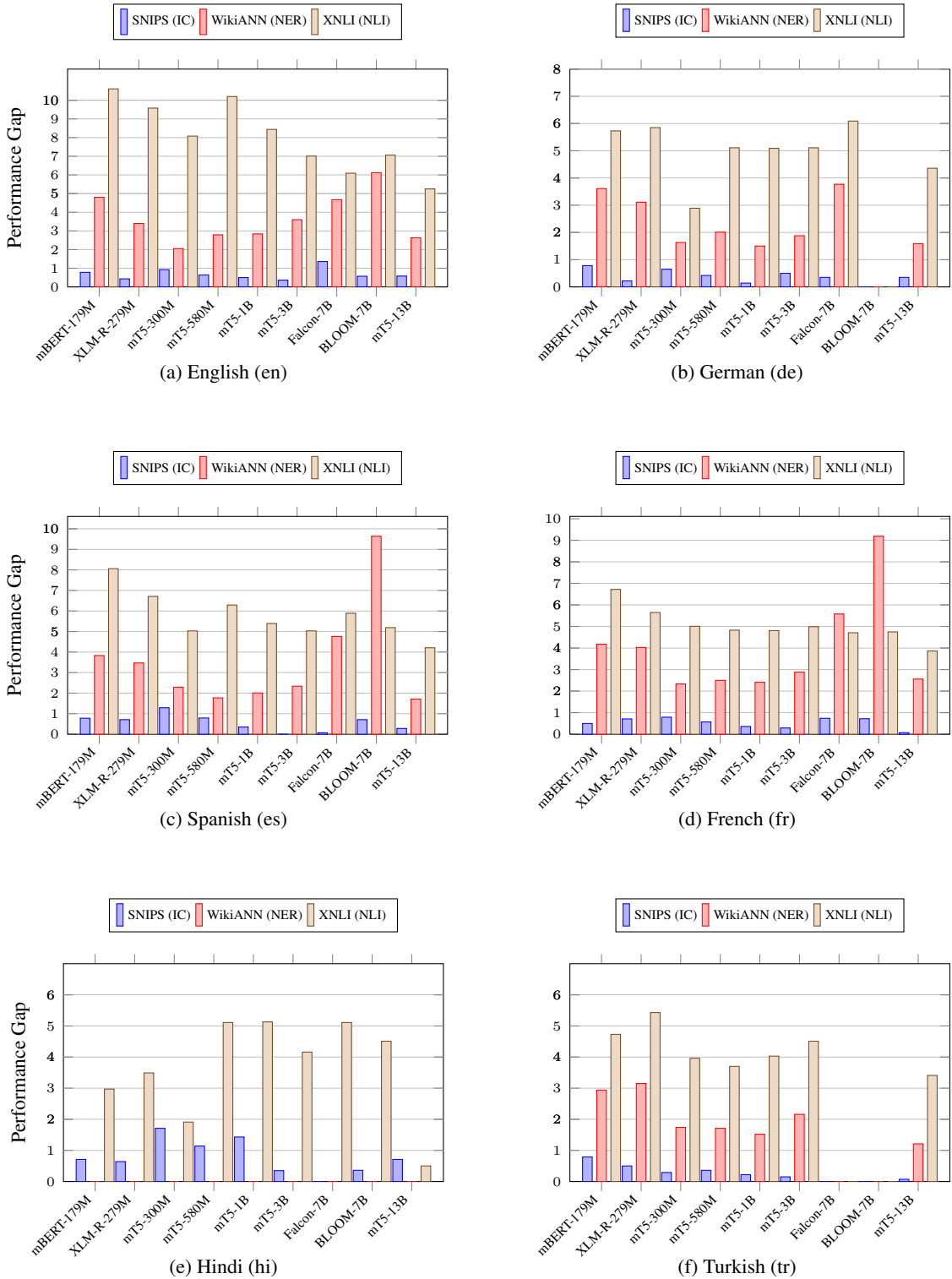
\begin{figure*}[t]
\centering
\captionsetup[subfigure]{skip=-5pt} 
\begin{subfigure}{0.45\textwidth}
\begin{tikzpicture}
\begin{axis}[
    ybar,
    width=7.5cm,
    height=5cm,
    bar width=0.15cm,
    xlabel={ },
    ylabel={Performance Gap},
    ylabel style = {font=\small}, 
    xtick=data,
    xticklabels={mBERT-179M, XLM-R-279M, mT5-300M, mT5-580M, mT5-1B, mT5-3B, Falcon-7B, BLOOM-7B, mT5-13B},
    legend style={at={(0.5,1.3)},
      anchor=north,legend columns=-1,font=\tiny},
    x tick label style={rotate=45, anchor=east, font=\tiny},
    y tick label style={font=\tiny},
    legend image code/.code={
        \draw [#1] (0cm,-0.2cm) rectangle (0.1cm,0.2cm); },
    ymin=0,
    ymajorgrids=true,
    grid style={line width=.1pt, draw=gray!10},
    major grid style={line width=.2pt,draw=gray!50},
    extra y ticks={1,2,3,4,5,6,7,8,9,10},  
    extra y tick style={grid=major}, 
    ]
\addplot coordinates {(1, 0.78) (2, 0.43) (3, 0.92) (4, 0.64) (5, 0.50) (6, 0.36) (7, 1.36) (8, 0.57) (9, 0.58)};
\addplot coordinates {(1, 4.80) (2, 3.40) (3, 2.05) (4, 2.79) (5, 2.84) (6, 3.60) (7, 4.67) (8, 6.12) (9, 2.63)};
\addplot coordinates {(1, 10.61) (2, 9.58) (3, 8.08) (4, 10.20) (5, 8.44) (6, 7.01) (7, 6.09) (8, 7.06) (9, 5.25)};

\legend{SNIPS (IC), WikiANN (NER), XNLI (NLI)}
\end{axis}
\end{tikzpicture}
\centering
\caption{English (en)}
\end{subfigure}
\hfill
\begin{subfigure}{0.45\textwidth}
\begin{tikzpicture}
\begin{axis}[
    ybar,
    width=7.5cm,
    height=5cm,
    bar width=0.15cm,
    xlabel={ },
    ylabel={},
    xtick=data,
    grid style={line width=.1pt, draw=gray!10},
    major grid style={line width=.2pt,draw=gray!50},
    xticklabels={mBERT-179M, XLM-R-279M, mT5-300M, mT5-580M, mT5-1B, mT5-3B, Falcon-7B, BLOOM-7B, mT5-13B},
    legend style={at={(0.5,1.3)},
    grid style={line width=.1pt, draw=gray!10},
    major grid style={line width=.2pt,draw=gray!50},
      anchor=north,legend columns=-1,font=\tiny},
    x tick label style={rotate=45, anchor=east, font=\tiny},
    y tick label style={font=\tiny},
    legend image code/.code={
        \draw [#1] (0cm,-0.15cm) rectangle (0.1cm,0.2cm); },
    extra y ticks={1,2,3,4,5,6,7,8},  
    extra y tick style={grid=major}, 
    ymin=0,
    ymax =8,
    extra y ticks={1,2,3,4,5,6,7,8,9,10},  
    extra y tick style={grid=major}, 
    ]
\addplot coordinates {(1, 0.78) (2, 0.22) (3, 0.65) (4, 0.42) (5, 0.14) (6, 0.50) (7, 0.35) (8, 0) (9, 0.35)};
\addplot coordinates {(1, 3.61) (2, 3.11) (3, 1.63) (4, 2.01) (5, 1.50) (6, 1.88) (7, 3.77) (8, 0) (9, 1.59)};
\addplot coordinates {(1, 5.73) (2, 5.85) (3, 2.89) (4, 5.11) (5, 5.09) (6, 5.11) (7, 6.09) (8, 0) (9,4.36)};

\legend{SNIPS (IC), WikiANN (NER), XNLI (NLI)}
\end{axis}
\end{tikzpicture}
\centering
\caption{German (de)}
\end{subfigure}
\hfill
\vspace{1cm}
\begin{subfigure}{0.45\textwidth}
\begin{tikzpicture}
\begin{axis}[
    ybar,
    width=7.5cm,
    height=5cm,
    bar width=0.15cm,
    xlabel={ },
    ylabel={Performance Gap},
    ylabel style = {font=\small}, 
    xtick=data,
    grid style={line width=.1pt, draw=gray!10},
    major grid style={line width=.2pt,draw=gray!50},
    xticklabels={mBERT-179M, XLM-R-279M, mT5-300M, mT5-580M, mT5-1B, mT5-3B, Falcon-7B, BLOOM-7B, mT5-13B},
    legend style={at={(0.5,1.3)},
      anchor=north,legend columns=-1,font=\tiny},
    x tick label style={rotate=45, anchor=east, font=\tiny},
    y tick label style={font=\tiny},
    legend image code/.code={
        \draw [#1] (0cm,-0.15cm) rectangle (0.1cm,0.2cm); },
    ymin=0,
    extra y ticks={1,2,3,4,5,6,7,8,9,10},  
    extra y tick style={grid=major}, 
    ]
\addplot coordinates {(1,0.78) (2, 0.71) (3, 1.29) (4, 0.79) (5, 0.35) (6, 0) (7, 0.07) (8, 0.71) (9, 0.28)};
\addplot coordinates {(1, 3.83) (2, 3.47) (3,2.29) (4, 1.77) (5, 2.01) (6, 2.34) (7, 4.76) (8, 9.64) (9, 1.71)};
\addplot coordinates {(1, 8.06) (2, 6.71) (3, 5.03) (4, 6.29) (5, 5.39) (6, 5.03) (7, 5.89) (8, 5.19) (9, 4.21)};

\legend{SNIPS (IC), WikiANN (NER), XNLI (NLI)}
\end{axis}
\end{tikzpicture}
\centering
\caption{Spanish (es)}
\end{subfigure}
\hfill
\begin{subfigure}{0.45\textwidth}
\begin{tikzpicture}
\begin{axis}[
    ybar,
    width=7.5cm,
    height=5cm,
    bar width=0.15cm,
    xlabel={ },
    xtick=data,
    grid style={line width=.1pt, draw=gray!10},
    major grid style={line width=.2pt,draw=gray!50},
    xticklabels={mBERT-179M, XLM-R-279M, mT5-300M, mT5-580M, mT5-1B, mT5-3B, Falcon-7B, BLOOM-7B, mT5-13B},
    legend style={at={(0.5,1.3)},
      anchor=north,legend columns=-1,font=\tiny},
    x tick label style={rotate=45, anchor=east, font=\tiny},
    y tick label style={font=\tiny},
    legend image code/.code={
        \draw [#1] (0cm,-0.15cm) rectangle (0.1cm,0.2cm); },
    ymin=0,
    extra y ticks={1,2,3,4,5,6,7,8,9},  
    extra y tick style={grid=major}, 
    ]
\addplot coordinates {(1, 0.50) (2, 0.71) (3, 0.79) (4, 0.57) (5, 0.36) (6, 0.29) (7, 0.74) (8, 0.72) (9, 0.07)};
\addplot coordinates {(1, 4.18) (2, 4.03) (3, 2.33) (4, 2.50) (5, 2.41) (6, 2.88) (7, 5.58) (8, 9.19) (9, 2.56)};
\addplot coordinates {(1, 6.72) (2, 5.65) (3, 5.01) (4, 4.83) (5, 4.81) (6, 4.99) (7, 4.71) (8, 4.75) (9, 3.86)};

\legend{SNIPS (IC), WikiANN (NER), XNLI (NLI)}
\end{axis}
\end{tikzpicture}
\centering
\caption{French (fr)}
\end{subfigure}
\hfill
\vspace{1cm}
\begin{subfigure}{0.45\textwidth}
\begin{tikzpicture}
\begin{axis}[
    ybar,
    width=7.5cm,
    height=5cm,
    bar width=0.15cm,
    xlabel={ },
    ylabel={Performance Gap},
    ylabel style = {font=\small}, 
    xtick=data,
    grid style={line width=.1pt, draw=gray!10},
    major grid style={line width=.2pt,draw=gray!50},
    xticklabels={mBERT-179M, XLM-R-279M, mT5-300M, mT5-580M, mT5-1B, mT5-3B, Falcon-7B, BLOOM-7B, mT5-13B},
    legend style={at={(0.5,1.3)},
      anchor=north,legend columns=-1,font=\tiny},
    x tick label style={rotate=45, anchor=east, font=\tiny},
    y tick label style={font=\tiny},
    legend image code/.code={
        \draw [#1] (0cm,-0.15cm) rectangle (0.1cm,0.2cm); },
    ymin=0,
    ymax= 7,
    extra y ticks={1,2,3,4,5,6},  
    extra y tick style={grid=major}, 
    ]
\addplot coordinates {(1, 0.71) (2, 0.64) (3, 1.71) (4, 1.14) (5, 1.43) (6, 0.35) (7, 0) (8, 0.36) (9, 0.71)};
\addplot coordinates {(1, 0) (2, 0) (3, 0) (4,0) (5, 0) (6, 0) (7, 0) (8, 0) (9, 0)};
\addplot coordinates {(1, 2.97) (2, 3.49) (3, 1.91) (4, 5.11) (5, 5.13) (6, 4.16) (7, 5.11) (8, 4.51) (9, 0.5)};

\legend{SNIPS (IC), WikiANN (NER), XNLI (NLI)}
\end{axis}
\end{tikzpicture}
\centering
\caption{Hindi (hi)}
\end{subfigure}
\hfill
\begin{subfigure}{0.45\textwidth}
\begin{tikzpicture}
\begin{axis}[
    ybar,
    width=7.5cm,
    height=5cm,
    bar width=0.15cm,
    xlabel={ },
    ylabel={},
    xtick=data,
    grid style={line width=.1pt, draw=gray!10},
    major grid style={line width=.2pt,draw=gray!50},
    xticklabels={mBERT-179M, XLM-R-279M, mT5-300M, mT5-580M, mT5-1B, mT5-3B, Falcon-7B, BLOOM-7B, mT5-13B},
    legend style={at={(0.5,1.3)},
      anchor=north,legend columns=-1,font=\tiny},
    x tick label style={rotate=45, anchor=east, font=\tiny},
    y tick label style={font=\tiny},
    legend image code/.code={
        \draw [#1] (0cm,-0.15cm) rectangle (0.1cm,0.2cm); },
    ymin=0,
    ymax = 7,
    extra y ticks={1,2,3,4,5,6},  
    extra y tick style={grid=major}, 
    ]
\addplot coordinates {(1, 0.79) (2, 0.50) (3, 0.29) (4, 0.36) (5, 0.22) (6, 0.15) (7, 0) (8, 0) (9, 0.07)};
\addplot coordinates {(1, 2.94) (2, 3.15) (3,1.74) (4,1.71) (5, 1.52) (6, 2.16) (7, 0) (8, 0) (9, 1.21)};
\addplot coordinates {(1, 4.73) (2, 5.43) (3, 3.96) (4, 3.70) (5, 4.03) (6, 4.51) (7, 0) (8, 0) (9, 3.41)};

\legend{SNIPS (IC), WikiANN (NER), XNLI (NLI)}
\end{axis}
\end{tikzpicture}
\centering
\caption{Turkish (tr)}
\end{subfigure}
\hfill
\vspace{1cm}

\caption{Performance gap between clean and noisy test sets of SNIPS (IC), XNLI (NLI) and WikiANN (NER) datasets for English (en), German (de), Spanish (es), French (fr), Hindi (hi), and Turkish (tr) languages. }
\label{all_languages_performance}
\end{figure*}

\begin{table*}[t]
    \centering
    \small
    \setlength{\tabcolsep}{4pt}
    
    \begin{tabular}{|p{5cm} |c| c| c | c|}  
     \hline
        \textbf{Model} & \textbf{Lr} & \textbf{Batch size} & \textbf{Weight decay} & \textbf{Epochs}  \\
        \hline
        \multicolumn{5}{|c|}{SNIPS} \\
        \hline
        mBERT, XLM-R, mT5-3B, Falcon-7B, BLOOM-7B, mT5-13B & 1e-05 & 8 & 0.1 & 2  \\
        mT5-300M, mT5-580M & 1e-04 & 8 & 0.01 & 6  \\

        mT5-1B &  1e-04 & 8 & 0.01 & 2  \\

        \hline
        \hline
        
        \multicolumn{5}{|c|}{WikiANN} \\
        \hline

        All Models & 3e-04 & 32 & 0.01 & 2  \\
        \hline
        \hline
        
                \multicolumn{5}{|c|}{XNLI} \\
       \hline
        
        mBERT, XLM-R, mT5-3B, Falcon-7B, BLOOM-7B, mT5-13B & 1e-05 & 32 & 0.01 & 2  \\
        mT5-300M, mT5-580M , mT5-1B & 1e-04 & 32 & 0.01 & 6  \\

        \hline
    \end{tabular}

    \caption{Hyper-parameters use to fine tune the models for each dataset. Other parameters are the same for all the experiments:
    gradient\_accumulation=4,
    optimizer=Adam,
    bf16=True,
    gradient\_checkpointing=True.}
    \label{tab:hyperparams}

\end{table*}



\end{document}